\documentclass[11pt]{article}
\usepackage{coling2020}
\usepackage{times}
\usepackage{url}
\usepackage{latexsym}
\usepackage{subcaption}
\usepackage{graphicx}
\usepackage{bm}

\usepackage{collcell}
\usepackage{pgf}
\usepackage{multirow}
\usepackage{booktabs}
\usepackage{hyperref}

\usepackage{colortbl}
\usepackage{pgfplots}
\usepackage{pgfplotstable}
\usepackage{ifthen}
\usepackage[normalem]{ulem}

\definecolor{myblue}{rgb}{0.9, 0.1, 0.94}
\definecolor{mygreen}{rgb}{0.64, 0.76, 0.68}
\definecolor{myyellow}{rgb}{0.88, 0.54, 0.35}
\definecolor{mygreen}{rgb}{0.68, 0.85, 0.9}

\def\colorModel{hsb} 

\newcommand\ColCell[1]{
  \ifthenelse{\equal{#1}{}}{%
  }{%
  \pgfmathparse{#1<50?1:0}  
  \ifnum\pgfmathresult=0\relax\color{white}\fi
  \pgfmathsetmacro\compA{0}      
  \pgfmathsetmacro\compB{(#1-0.3)} 
  \pgfmathsetmacro\compC{1}      
  \edef\x{\noexpand\centering\noexpand\cellcolor[\colorModel]{\compA,\compB,\compC}}\x{#1}
  }}
\newcolumntype{E}{>{\collectcell\ColCell}m{0.85cm}<{\endcollectcell}} 
\colingfinalcopy 

\title{Metrics also Disagree in the Low Scoring Range: Revisiting \\ Summarization Evaluation Metrics}

\author{Manik Bhandari \qquad
  Pranav Gour \qquad
  Atabak Ashfaq \qquad
  Pengfei Liu \qquad \\
    Carnegie Mellon University \\
    \{\texttt{mbhandar,pgour,aashfaq,pliu3\}@cs.cmu.edu}
  }

\date{}

\begin{document}
\maketitle
\begin{abstract}

In text summarization, evaluating the efficacy of automatic metrics without human judgments has become recently popular.
One exemplar work \cite{peyrard-2019-studying} concludes that automatic metrics strongly disagree when ranking high-scoring summaries. In this paper, we revisit their experiments and find that their observations stem from the fact that \emph{metrics disagree in ranking summaries from any narrow scoring range}. We hypothesize that this may be because summaries are similar to each other in a narrow scoring range and are thus, difficult to rank. Apart from the width of the scoring range of summaries, we analyze three other properties that impact inter-metric agreement - \emph{Ease of Summarization}, \emph{Abstractiveness}, and \emph{Coverage}. To encourage reproducible research, we make all our analysis code and data publicly available.\footnote{\href{https://github.com/manikbhandari/RevisitSummEvalMetrics}{https://github.com/manikbhandari/RevisitSummEvalMetrics}}


\end{abstract}

\section{Introduction}
\label{intro}

%
%

Automatic metrics play a significant role in summarization evaluation, profoundly affecting the direction of system optimization.
Due to its importance, evaluating the quality of evaluation metrics, also known as \emph{meta-evaluation} has been a crucial step.
Generally, there are two meta-evaluation strategies: (i) assessing how well each metric correlates with human judgments ~\cite{lin2004rouge,ng-abrecht-2015-better,js2,peyrard_s3,Bhandari-2020-reevaluating}, which requires procuring manual annotations that are expensive and time-consuming, and (ii) measuring the correlation between different metrics \cite{peyrard-2019-studying}, which is a human judgment-free method.
In this work, we focus on the latter and 
ask two research questions:

\paragraph{RQ1:}  How do automated metrics correlate when ranking summaries in different scoring ranges (low, average, and high)? 
We revisit the experiments of~\newcite{peyrard-2019-studying} which concludes that automated metrics strongly disagree for ranking high-scoring summaries.~\footnote{\newcite{peyrard-2019-studying} uses three experiments to reach their conclusion. Due to limitations of space, we focus on the first one here. Please see the appendix for a detailed analysis of the other two experiments.} We find that the scoring range has little effect on the correlation of metrics. It is rather the \emph{width of the scoring range} which affects inter-metric correlation. Specifically, we observe that metrics agree in ranking summaries from the full scoring range but disagree in ranking summaries from low, average, and high scoring ranges when taken separately.

\blfootnote{
    %
    %
    %
    \hspace{-0.65cm}  
    This work is licensed under a Creative Commons 
    Attribution 4.0 International License.
    License details:
    \url{http://creativecommons.org/licenses/by/4.0/}.
}

\paragraph{RQ2:} Which other factors affect the correlations of metrics?
In addition to the width of the scoring range, we analyze three properties of a reference summary on inter-metric correlation
- \emph{Ease of Summarization}, \emph{Abstractiveness} and \emph{Coverage}. Overall we find that for highly extractive document-reference summary pairs, inter-metric correlation is high whereas metrics disagree when ranking summaries of abstractive document-reference summary pairs.\\


\noindent
We summarize our contributions as follows:
(1) We extend the analysis of \newcite{peyrard-2019-studying} and find that not only do metrics disagree in the high scoring range, they also disagree in the low and medium scoring range.
(2) We perform our analysis on the popular CNN/Dailymail dataset using traditional lexical matching metrics like ROUGE as well as recently popular semantic matching metrics like BERTScore and MoverScore.
(3) Apart from the width of the scoring range, we analyze three linguistic properties of reference summaries which affect inter-metric correlations.

\section{Preliminaries}  \label{sec:hj-free}

\subsection{Datasets}
\noindent\textbf{TAC-2008, 2009}~\cite{tac2008,tac2009} are multi-document, multi-reference summarization datasets used during the TAC-2008, TAC-2009 shared tasks. Following~\cite{peyrard-2019-studying} we combine the two and refer to the joined dataset as \texttt{TAC}.

\noindent\textbf{CNN/DailyMail (CNNDM)} \cite{hermann2015teaching} is a commonly used summarization dataset modified by \newcite{nallapati2016abstractive}, which contains news articles and
associated highlights as summaries. We use the non-anonymized version.


\subsection{Evaluation Metrics}
\label{sec:metrics}
We examine six metrics that measure the semantic equivalence between two texts, in our case, between the system-generated summary and the reference summary.  
\textbf{BERTScore (BScore)} measures soft overlap between contextual BERT embeddings of tokens between the two texts\footnote{BERTScore:  \href{https://github.com/Tiiiger/bert_score}{github.com/Tiiiger/bert\_score}} \cite{bert-score}.
\textbf{MoverScore (MS)} applies a distance measure to contextualized BERT and ELMo word embeddings\footnote{MoverScore:  \href{https://github.com/AIPHES/emnlp19-moverscore}{github.com/AIPHES/emnlp19-moverscore}} \cite{zhao-etal-2019-moverscore}.
\textbf{JS divergence (JS-2)} measures Jensen-Shannon divergence between the two text's bigram distributions\footnote{JS-2: the function defined in \href{https://github.com/UKPLab/coling2016-genetic-swarm-MDS}{github.com/UKPLab/coling2016-genetic-swarm-MDS}} \cite{lin-etal-2006-information}. 
\textbf{ROUGE-1 (R1)} and \textbf{ROUGE-2 (R2)} measure the overlap of unigrams and bigrams respectively\footnote{\label{pyrouge} ROUGE-1,2, and L: the python wrapper: \href{https://github.com/sebastianGehrmann/rouge-baselines}{github.com/sebastianGehrmann/rouge-baselines}} \cite{lin2004rouge}.
\textbf{ROUGE-L} measures the overlap of the longest common subsequence between two texts \cite{lin2004rouge}.
We use the recall variant of all metrics except MoverScore which has no specific recall variant.

\subsection{Correlation Measure}
\noindent\textbf{Kendall's} \bm{$\tau$ }
is a measure of the rank correlation between any two measured quantities (in our case scores given by evaluation metrics) and is popular in meta-evaluating metrics at the summary level \cite{peyrard-2019-studying}. We use the implementation given by~\newcite{scipy-2020}.


\section{Summary Generation}
To simulate the full scoring range of summaries that are possible for a document, we follow \newcite{peyrard-2019-studying} and use a genetic algorithm \cite{peyrard_genetic} to generate extractive summaries. We optimize for 5 metrics - ROUGE-1, ROUGE-2, ROUGE-L, BERTScore, and MoverScore, generating 100 summaries per metric for each of the nearly 11K documents in the \texttt{CNNDM} test set resulting in 500 summaries per document. After de-duplication, we are left with nearly 419 summaries per document on average. For the \texttt{TAC} dataset, we randomly sample 500 summaries for each document from the nearly 2000 output summaries provided by \newcite{peyrard-2019-studying}. 

\section{Experiment and Analysis}
\subsection{Width of Scoring Range}
\label{sec:hj_free_summ}


In this experiment, we aim to re-examine the results in~\newcite{peyrard-2019-studying} and answer our first research question \textbf{Q1}: how do different automated metrics correlate in ranking summaries in different scoring ranges?
We approach this as follows: for each summary $s_{ij}$ of document $d_i$, we first calculate its mean score across all metrics after normalizing the metrics to be between $0$ and $1$.
We use this to partition the scoring range of each document into three parts: low scoring (L), medium scoring (M), and top scoring (T), which are the bottom third, the middle third and the top third of the scoring range respectively. We then analyze the summaries falling into these bins in two different ways:

\noindent
\textbf{1. Cumulative:} In this setting we aim to replicate \newcite{peyrard-2019-studying}'s results, which compared inter-metric agreement on the whole set of summaries to that on the top-scoring subset. To do this, we compute the average inter-metric correlation for summaries belonging to (i) L + M + T, (ii) M + T and (iii) T as shown in the left side in Tab.~(\ref{tab:ktau_tac}-\ref{tab:ktau_cnndm}).
Note that here the width of the scoring range is different for each row.

\noindent
\textbf{2. Non-cumulative:} In this setting, we analyze the average inter-metric correlation on summaries belonging to each scoring bin separately as shown in the right side of Tab.~(\ref{tab:ktau_tac}-\ref{tab:ktau_cnndm}). We advocate for the use of this setting as (1) it controls for the width of the scoring range and (2)   it allows for a more fine-grained analysis of the scoring range.

\noindent Note that, for each bin, the correlation is calculated for summaries generated for each document and then averaged over all documents. We only consider statistically significant ($p < 0.05$) kendall's $\tau$ values.

\noindent
\textbf{Observations \& Discussion} 
Our observations on the \texttt{TAC} and \texttt{CNNDM} datasets are shown in Tab.~\ref{tab:ktau_tac} and \ref{tab:ktau_cnndm} respectively.
In the cumulative setting, we observe the same trend reported by ~\newcite{peyrard-2019-studying}: inter-metric agreement decreases when the average score increases and is the lowest in the top scoring range (T). 
However, in the non-cumulative setting, where metrics rank summaries from a narrow scoring range, we observe that (i) metrics have low correlations in all three scoring ranges (low, medium, and top) and (ii) there is no clear trend in correlations across the bins. Comparing the cumulative and non-cumulative settings, one can see that decreasing the width of the scoring range reduces the inter-metric correlations.
This suggests that rather than the scoring range, the width of the scoring range has a strong impact on the correlation between metrics. 
This may be because summaries from a narrow scoring range are similar to each other, and thus, difficult for different metrics to rank consistently.
\begin{table}
    \centering \footnotesize
    \newcommand\items{7}   
    \resizebox{0.5\textwidth}{!}{
    \begin{tabular}{llEEEEE}
        \toprule  
        Metric & Bin & \multicolumn{1}{c}{MS} & \multicolumn{1}{c}{R1} & \multicolumn{1}{c}{R2} & \multicolumn{1}{c}{RL} & \multicolumn{1}{c}{JS2} \\
        \midrule
        & L+M+T  & 0.66 & 0.57 & 0.66 & 0.58 & 0.64 \\
        {BScore} & M+T & 0.60 & 0.49 & 0.60 & 0.51 & 0.59 \\
        & T & 0.33 & 0.29 & 0.33 & 0.29 & 0.32 \\
        \midrule
        & L+M+T    &  & 0.65 & 0.69 & 0.66 & 0.69 \\
        {MS} & M+T &  & 0.58 & 0.63 & 0.59 & 0.63 \\
        & T        &  & 0.38 & 0.35 & 0.38 & 0.35 \\
        \midrule
        & L+M+T  &  &  & 0.67 & 0.92 & 0.59 \\
        {R1} & M+T &  &  & 0.61 & 0.90 & 0.51 \\
        & T &  &  & 0.46 & 0.85 & 0.30 \\
        \midrule
        & L+M+T  &  &  &  & 0.69 & 0.83 \\
        {R2} & M+T &  &  &  & 0.64 & 0.80 \\
        & T &  &  &  & 0.49 & 0.61 \\
        \midrule
        & L+M+T  &  &  &  &  & 0.61 \\
        {RL} & M+T &  &  &  &  & 0.54 \\
        & T &  &  &  &  & 0.32 \\
        \bottomrule
    \end{tabular}
    }
    \resizebox{0.397\textwidth}{!}{
    \begin{tabular}{c*{\items}{E}}
        \toprule  
        Bin & \multicolumn{1}{c}{MS} & \multicolumn{1}{c}{R1} & \multicolumn{1}{c}{R2} & \multicolumn{1}{c}{RL} & \multicolumn{1}{c}{JS2} \\
        \midrule
        L   & 0.30 & 0.09 & 0.24 & 0.08 & 0.22 \\
        M  & 0.40 & 0.21 & 0.38 & 0.21 & 0.37 \\
        T & 0.33 & 0.29 & 0.33 & 0.29 & 0.32 \\
        \midrule
        L   &  & 0.26 & 0.30 & 0.24 & 0.31 \\
        M  &  & 0.29 & 0.38 & 0.29 & 0.41 \\
        T &  & 0.38 & 0.35 & 0.38 & 0.35 \\
        \midrule
        L   &  &  & 0.28 & 0.83 & 0.11 \\
        M  &  &  & 0.28 & 0.82 & 0.19 \\
        T &  &  & 0.46 & 0.85 & 0.30 \\
        \midrule
        L   &  &  &  & 0.29 & 0.57 \\
        M  &  &  &  & 0.32 & 0.69 \\
        T &  &  &  & 0.49 & 0.61 \\
        \midrule
        L   &  &  &  &  & 0.14 \\
        M  &  &  &  &  & 0.22 \\
        T &  &  &  &  & 0.32 \\
        \bottomrule
    \end{tabular}
    }
    \caption{\label{tab:ktau_tac}Kendall's $\tau$ for the cumulative and non-cumulative settings on \texttt{TAC}. Higher values indicate greater inter-metric correlation. L, M and T correspond to the low, medium and top scoring range respectively. Please see Sec.~\ref{sec:hj_free_summ} for more details.
    }
\end{table}
\begin{table}
    \centering \footnotesize
    \newcommand\items{7}   
    \resizebox{0.5\textwidth}{!}{
    \begin{tabular}{ll*{\items}{E}}
        \toprule  
        Metric & \multicolumn{1}{l}{Bin} & \multicolumn{1}{c}{MS} & \multicolumn{1}{c}{R1} & \multicolumn{1}{c}{R2} & \multicolumn{1}{c}{RL} & \multicolumn{1}{c}{JS2} \\
        \midrule
        & L+M+T & 0.75 & 0.72 & 0.7 & 0.71 & 0.64 \\
        {BScore} & M+T & 0.62 & 0.61 & 0.59 & 0.6 & 0.5 \\
        & T & 0.39 & 0.4 & 0.36 & 0.39 & 0.19 \\
        \midrule
        & L+M+T  & & 0.69 & 0.68 & 0.68 & 0.67 \\
        {MS} & M+T & & 0.53 & 0.55 & 0.52 & 0.59 \\
        & T &  & 0.23 & 0.24 & 0.21 & 0.37 \\
        \midrule
        & L+M+T  &  &  & 0.75 & 0.91 & 0.66 \\
        {R1} & M+T &  &  & 0.69 & 0.88 & 0.5  \\
        & T &  &  & 0.53 & 0.82 & 0.16 \\
        \midrule
        & L+M+T  &  &  &  & 0.77 & 0.86 \\
        {R2} & M+T  &  &  &  & 0.72 & 0.76 \\
        & T &  &  &  & 0.56 & 0.56 \\
        \midrule
        & L+M+T  &  &  &  &  & 0.67 \\
        {RL} & M+T  &  &  &  &  & 0.53 \\
        & T &  &  &  &  & 0.2 \\
        \bottomrule
    \end{tabular}
    }
    \resizebox{0.397\textwidth}{!}{
    \begin{tabular}{l*{\items}{E}}
        \toprule  
        Bin & \multicolumn{1}{c}{MS} & \multicolumn{1}{c}{R1} & \multicolumn{1}{c}{R2} & \multicolumn{1}{c}{RL} & \multicolumn{1}{c}{JS2} \\
        \midrule
        L  & 0.49 & 0.38 & 0.32 & 0.36 & 0.27 \\
        M & 0.51 & 0.45 & 0.41 & 0.43 & 0.37 \\
        T & 0.39 & 0.4 & 0.36 & 0.39 & 0.19 \\
        \midrule
        L  & & 0.35 & 0.29 & 0.33 & 0.28 \\
        M &  & 0.4  & 0.39 & 0.39 & 0.4 \\
        T &  & 0.23 & 0.24 & 0.21 & 0.37 \\
        \midrule
        L  &  &  & 0.38 & 0.78 & 0.28 \\
        M &  &  & 0.47 & 0.78 & 0.38 \\
        T &  &  & 0.53 & 0.82 & 0.16 \\
        \midrule
        L  &  &  &  & 0.4  & 0.69 \\
        M &  &  &  & 0.52 & 0.78  \\
        T &  &  &  & 0.56 & 0.56 \\
        \midrule
        L   &  &  &  &  & 0.3 \\
        M  &  &  &  &  & 0.41 \\
        T &  &  &  &  & 0.2 \\
        \bottomrule
    \end{tabular}
    }
    \caption{\label{tab:ktau_cnndm}Kendall's $\tau$ for the cumulative and non-cumulative settings on \texttt{CNNDM}. Higher values indicate greater inter-metric correlation. L, M and T correspond to the low, medium and top scoring range respectively. Please see Sec.~\ref{sec:hj_free_summ} for more details.
    }
\end{table}



\subsection{Factors affecting Inter-metric Correlation}
\label{sec:sys_level_metrics}
In this experiment, we aim to answer the second research question \textbf{Q2}: Apart from the width of the scoring range, which factors affect inter-metric correlations?
Specifically, we identify three factors which affect the correlation of metrics - (1) Ease of Summarization, (2) Abstractiveness, and (3) Coverage.

\noindent\textbf{1. Ease of Summarization (EoS): }
For each generated summary $s_{ij}$ of document $d_i$ with reference summary $r_i$, we define EoS as
$
    \mathrm{EoS}(d_i) = \frac{1}{n}\sum_{k = 1}^n \left[\max_j m_k(s_{ij}, r_i) \right].
$
Here, $m_k$ is a metric function normalized to be between $0$ and $1$.
Thus, EoS is the average over all metrics of the maximum score that any summary received. A higher EoS score for a document implies that for that document, we can generate higher scoring extractive summaries according to many metrics.

\noindent\textbf{2. Abstractiveness: }
We define abstracriveness of a document $d_i$ with reference $r_i$ as $\frac{1 - |\mathrm{Voc}(d_i) \cap \mathrm{Voc}(r_i)|}{|\mathrm{Voc}(r_i)|}$ where $\mathrm{Voc}(x)$ is the set of unique tokens of any text $x$.
Abstractiveness measures the overlap in vocabularies of the document and its reference summary.

\noindent \textbf{3. Coverage: }
We use the definition of Coverage as provided by \newcite{grusky2018newsroom} i.e. ``the percentage of words in the summary that are part of an extractive fragment with the article". We refer the reader to \newcite{grusky2018newsroom} for a detailed description of Coverage.

\begin{figure}
    \centering
    \subfloat[\texttt{TAC} EoS]{
    \includegraphics[width=0.1825\linewidth]{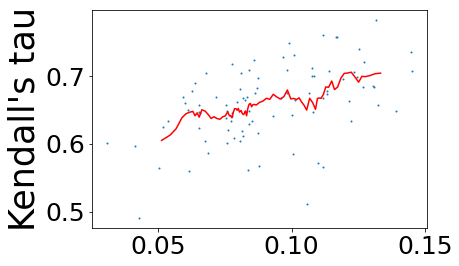}
    }
    \subfloat[\texttt{TAC} Abs]{
    \includegraphics[width=0.1475\linewidth]{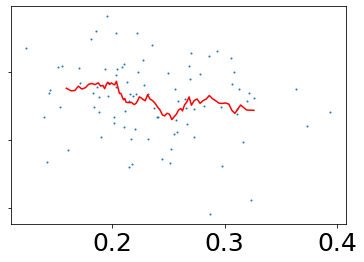}
    }
    \subfloat[\texttt{TAC} Cov]{
    \includegraphics[width=0.15\linewidth]{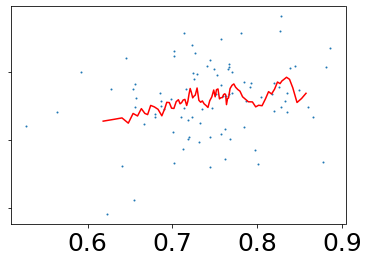}
    }
    \subfloat[\texttt{CNNDM} EoS]{
    \includegraphics[width=0.165\linewidth]{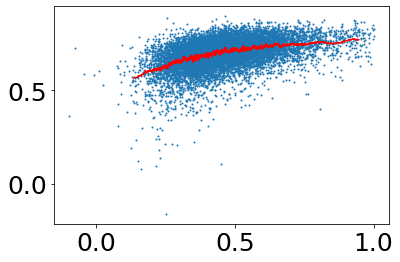}
    }
    \subfloat[\texttt{CNNDM} Abs]{
    \includegraphics[width=0.15\linewidth]{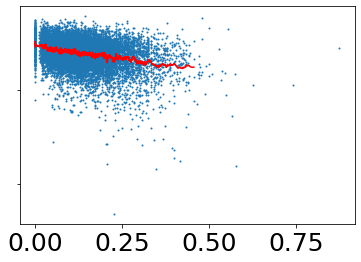}
    }
    \subfloat[\texttt{CNNDM} Cov]{
    \includegraphics[width=0.15\linewidth]{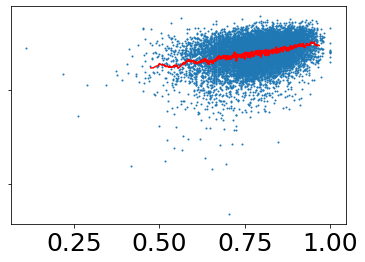}
    }
    \caption{Effect of different properties of reference summaries. 
    We only show correlation between BERTScore and ROUGE-2 due to limited pages.    
    The trend is similar for all other pairs as shown in the appendix. The plots for \texttt{CNNDM} are more dense because of more documents in the \texttt{CNNDM} test set as compared to \texttt{TAC}. ``Cov" and ``Abs" stand for Coverage and Abstractiveness respectively. The trend lines in red are the 10 point and 100 point moving average for \texttt{TAC} and \texttt{CNNDM} respectively.
    }
    \label{fig:q2}
\end{figure}

\noindent
\textbf{Observations:} 
Our observations are summarized in Fig.~\ref{fig:q2}. Each point in the graph represents a document-reference summary pair with its corresponding property on the x-axis and inter-metric correlation of its summaries on the y-axis.
We find that

(1) metrics agree with each other as documents become easier to summarize

(2) as documents become more abstractive, the correlation between metrics decreases

(3) as the coverage of documents increases, the correlation between metrics increases.

\noindent
These observations suggest that automatic evaluation metrics have higher correlations for easier to summarize, and more extractive (lower abstractiveness, higher coverage) document-reference summary pairs.

\section{Implications and Future Directions}

In this work, we revisit the conclusion of \newcite{peyrard-2019-studying}'s work and show that instead of solely disagreeing in high-scoring range, metrics disagree when ranking summaries from all three scoring ranges - low, medium and top. This highlights the need to collect human judgments to identify trustworthy metrics.
Moreover, future meta-evaluations should use uniform-width bins when comparing correlations to ensure a more robust analysis. 
Additionally, we analyze three linguistic properties of reference summaries and their effect on inter-metric correlations. 
Our observation that metrics de-correlate as references become more abstractive suggests that we need to exercise caution when using automatic metrics to compare summarization systems on abstractive datasets like \texttt{XSUM}~\cite{narayan2018don}.
Moreover, future work proposing new evaluation metrics can analyze them using these properties to get more insights about their behavior. 
\section{Acknowledgments}
We would like to thank Maxime Peyrard for sharing the code and data used in~\newcite{peyrard-2019-studying} and for his useful feedback about our experiments. We would also like to thank Graham Neubig for his feedback and for providing the computational resources needed for this work.




\bibliographystyle{coling}
\bibliography{coling2020}
\newpage

\appendix
\section{Disagreement}
In addition to Kendall's $\tau$ between metrics, \newcite{peyrard-2019-studying} analyzes the disagreement between metrics and shows higher inter-metric disagreement in the higher scoring range. To analyze disagreement, they randomly sample 100,000 pairs of summaries (say $s_{a}$ and $s_b$ with corresponding references $r_a$, $r_b$) for each pair of metrics (say $m_1$ and $m_2$) and bin them into 15 \emph{cumulative bins} according to the average score for any one metric i.e. according to $\frac{1}{2}(m_1(s_a, r_a) + m_1(s_b, r_b))$. The disagreement for each bin is then calculated as the percentage of summary pairs for which $m_1(s_a, r_a) > m_1(s_b, r_b)$ but $m_2(s_a, r_a) < m_2(s_b, r_b)$ or vice-versa i.e. $m_1(s_a, r_a) < m_1(s_b, r_b)$ but $m_2(s_a, r_a) > m_2(s_b, r_b)$.

\begin{figure}
    \centering
    \subfloat[Cumulative bins on \texttt{TAC}]{
    \includegraphics[width=0.5\linewidth]{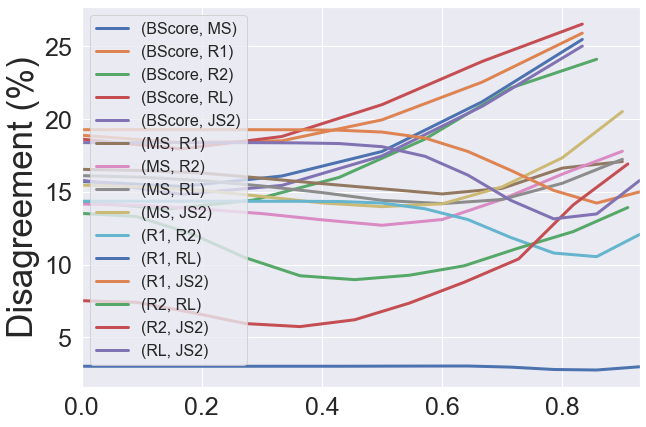}
    }
    \subfloat[Non-cumulative bins on \texttt{TAC}]{
    \includegraphics[width=0.475\linewidth]{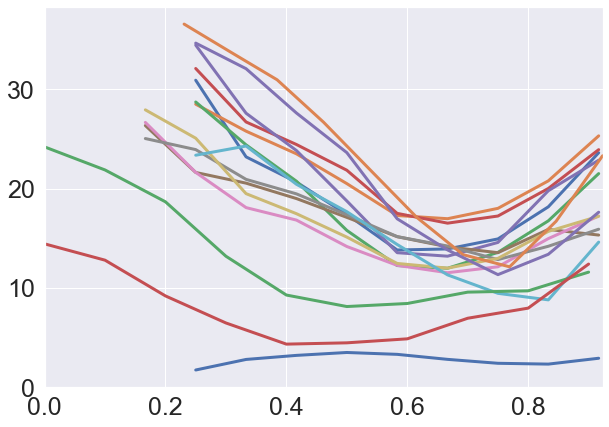}
    }

    \subfloat[Cumulative bins on \texttt{CNNDM}]{
    \includegraphics[width=0.5\linewidth]{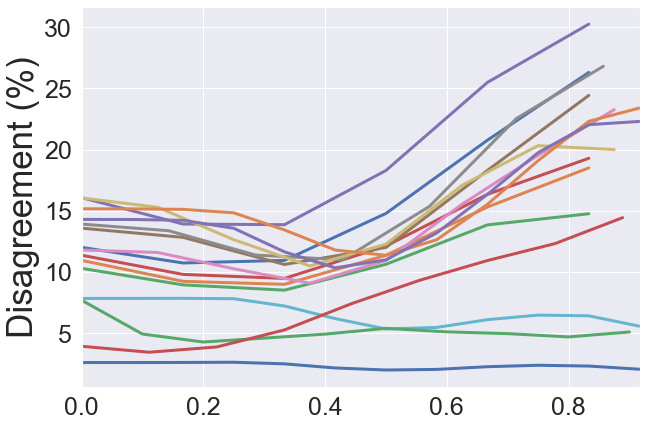}
    }
    \subfloat[Non-cumulative bins on \texttt{CNNDM}]{
    \includegraphics[width=0.475\linewidth]{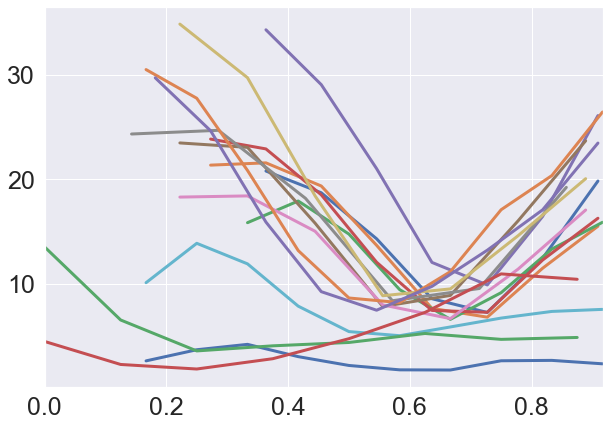}
    }
    \caption{Disagreement between metrics on \texttt{TAC} and \texttt{CNNDM}.}
    \label{fig:disagreement}
\end{figure}

The use of cumulative bins suffers from the same phenomena as described in section 4.1 i.e the width of the bin may play a role in the agreement of metrics. In Fig.~\ref{fig:disagreement} we replicate the cumulative disagreement plot for the \texttt{TAC} and \texttt{CNNDM} datasets and show the corresponding non-cumulative versions. We observe that when we control for the width of scoring range, inter-metric disagreement is higher even in the low scoring range.

\section{F/N Ratio}
\begin{figure}
    \centering
    \subfloat[\texttt{TAC}]{
    \includegraphics[width=0.33\linewidth]{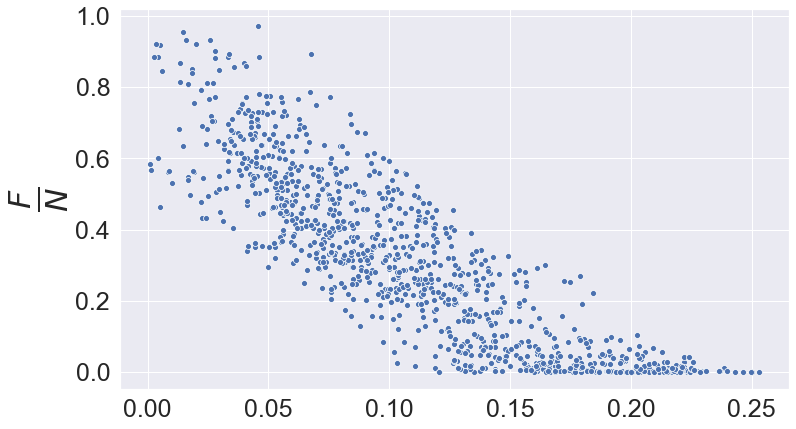}
    \label{fig:tac_f_n}
    }
    \subfloat[\texttt{CNNDM}]{
    \includegraphics[width=0.305\linewidth]{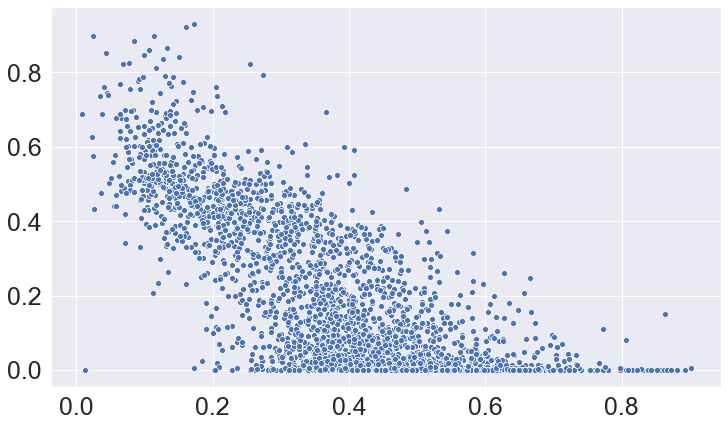}
    \label{fig:cnndm_f_n}
    }
    \subfloat[\texttt{CNNDM} with random metrics]{
    \includegraphics[width=0.305\linewidth]{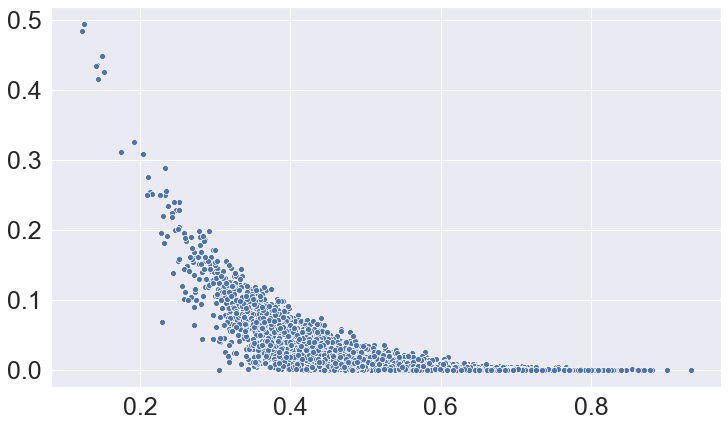}
    \label{fig:f_n_random}
    }
    
    \caption{F/N ratio between metrics on \texttt{TAC} and \texttt{CNNDM}.}
    \label{fig:f_n}
\end{figure}

\newcite{peyrard-2019-studying}'s final experiment measures if improvements according to one metric are consistent across other metrics. To this end they define $F/N$ as follows:

Let $S(d_i)$ be the set of summaries for a document $d_i$ and $M$ be the set of all metrics. Let's sample a summary $s \in S(d_i)$ randomly. Then

\begin{equation}
    \frac{F}{N} = \frac{|\{x \in S(d_i)| \forall m \in M, m(x) > m(s) \}|}{|\{x \in S(d_i)| \exists m \in M, m(x) > m(s) \}|}
\end{equation}
i.e. out of all the summaries ranked better than a summary by one metric, how many are ranked better by all the metrics. As shown in Fig.~\ref{fig:tac_f_n} on the \texttt{TAC} dataset, as the average score of $s$ (averaged across all metrics) increases, $F/N$ decreases. This may suggest that as summary quality improves, different metrics do not agree on which summaries are of better quality.

However, this quantity is misleading. As the average score of $s$ increases, the numerator $F$ will naturally decrease (because for a higher scoring $s$, the number of summaries that are better than $s$ are fewer) while the denominator $N$ may remain large even if one metric is misaligned with others. To prove this hypothesis, we first replicate the measure for \texttt{TAC} and \texttt{CNNDM} datasets in Fig.~\ref{fig:tac_f_n},~\ref{fig:cnndm_f_n}. Next, instead of the real metric scores, we assign each summary a random number sampled from $\mathrm{Uniform}(0, 1)$. In Fig.~\ref{fig:f_n_random} we see the same trend for random scores as for real metric scores. This shows that this decreasing trend is indeed a property of the ratio $F/N$ rather than being a property specific to real evaluation metrics. 

\begin{figure}
    \centering
    \subfloat[\texttt{TAC}]{
    \includegraphics[width=0.33\linewidth]{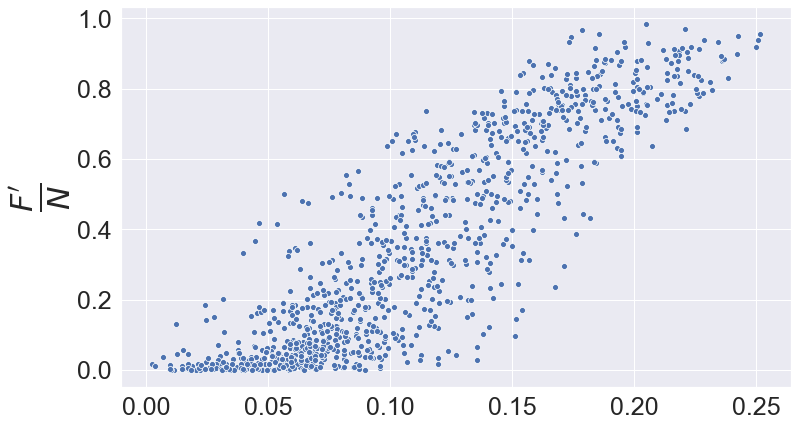}
    }
    \subfloat[\texttt{CNNDM}]{
    \includegraphics[width=0.305\linewidth]{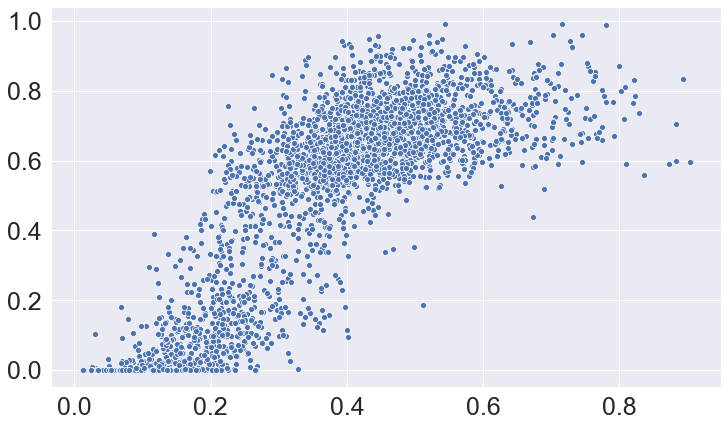}
    }
    \subfloat[\texttt{CNNDM} with random metrics]{
    \includegraphics[width=0.305\linewidth]{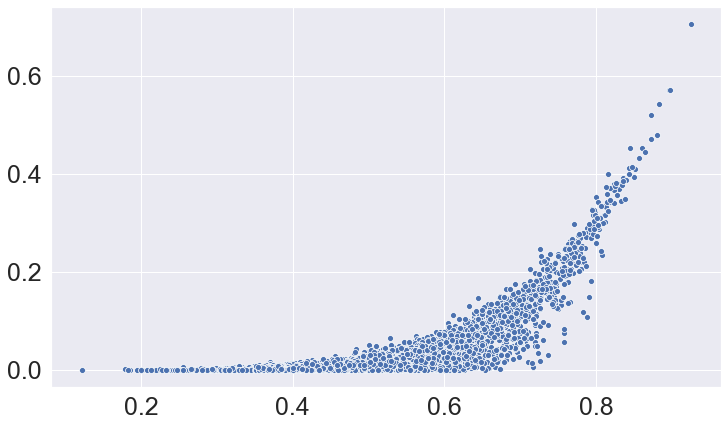}
    }
    
    \caption{$F'/N$ ration between metrics on \texttt{TAC} and \texttt{CNNDM}.}
    \label{fig:f_dash_n}
\end{figure}

Moreover, one can come up with a modified ratio $F'/N$ as follows

\begin{equation}
    \frac{F'}{N} = \frac{|\{x \in S(d_i)| \forall m \in M, m(x) < m(s) \}|}{|\{x \in S(d_i)| \exists m \in M, m(x) < m(s) \}|}
\end{equation}

which measures ``out of all the summaries that are ranked worse than a summary by one metric, how many are ranked worse by all metrics. If metrics truly de-correlated in only the higher scoring range, one would expect the same decreasing trend for $F'/N$. However, as is clear from Fig.~\ref{fig:f_dash_n} the trend is reversed for real as well as random metric scores. $F'/N$ increases as average score of $s$ increases. This is because, similar to $F/N$, this measure is also misleading and sensitive to the numerator $F'$ which always increases as average score of $s$ increases.

\section{Factors affecting inter-metric correlation}
\subsection{Ease of summarization}
\begin{figure}[ht]
    \centering
    \foreach \x in {BScore_MS, BScore_R1, BScore_R2, BScore_RL, BScore_JS2, MS_R1, MS_R2, MS_RL, MS_JS2, R1_R2, R1_RL, R1_JS2, R2_RL, R2_JS2, RL_JS2}
    {
        \includegraphics[scale=0.2]{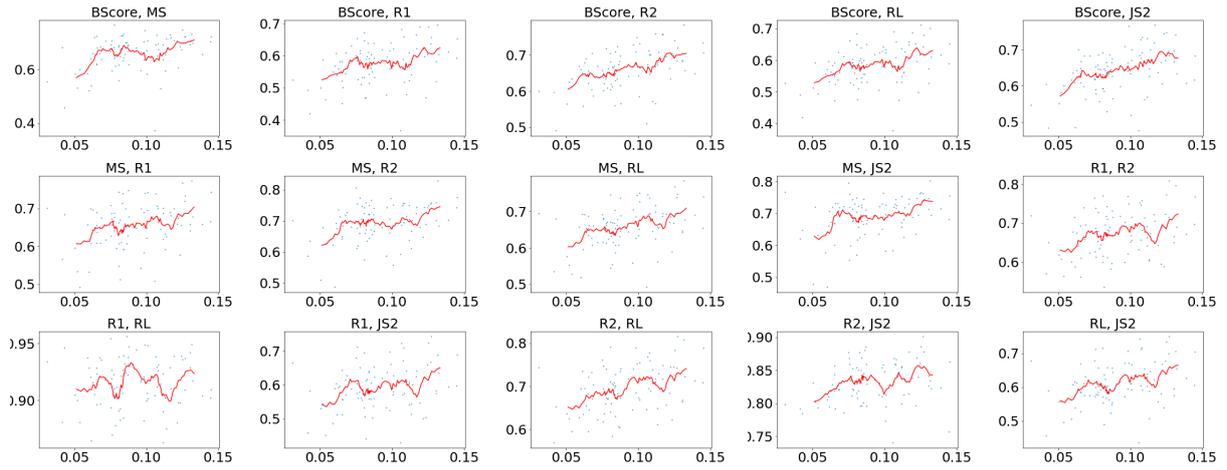}
    }    
    \caption{Ease of Summarization (x-axis) vs Kendall's $\tau$ (y-axis) for all metric pairs on the \texttt{TAC} dataset.}
    \label{fig:tac_eos_all_pairs}
\end{figure}

\begin{figure}[ht]
    \centering
    \foreach \x in {BScore_MS, BScore_R1, BScore_R2, BScore_RL, BScore_JS2, MS_R1, MS_R2, MS_RL, MS_JS2, R1_R2, R1_RL, R1_JS2, R2_RL, R2_JS2, RL_JS2}
    {
        \includegraphics[scale=0.2]{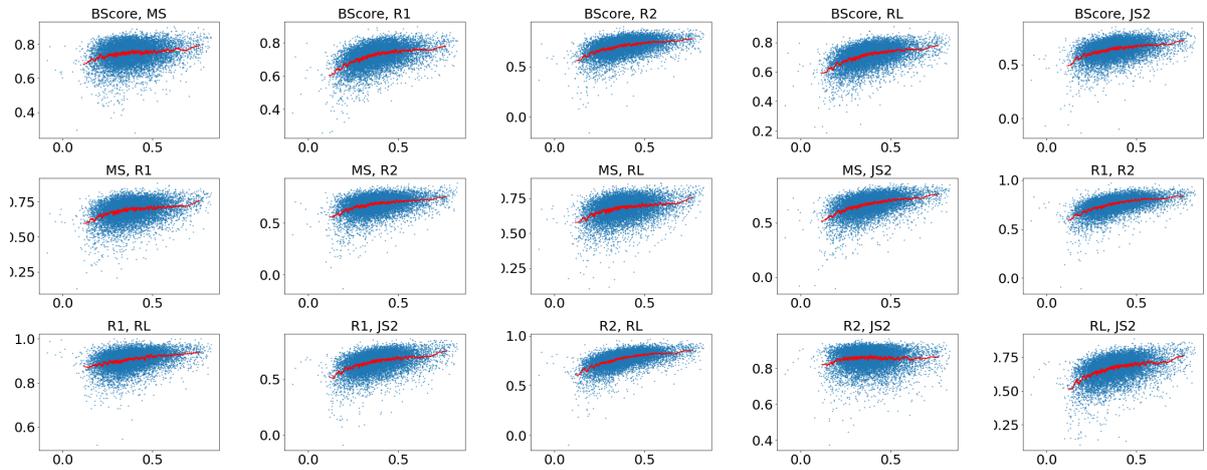}
    }    
    \caption{Ease of Summarization (x-axis) vs Kendall's $\tau$ (y-axis) for all metric pairs on the \texttt{CNNDM} dataset.}
    \label{fig:cnndm_eos_all_pairs}
\end{figure}

Please see Fig.~\ref{fig:tac_eos_all_pairs},~\ref{fig:cnndm_eos_all_pairs} for Ease of Summarization vs Kendall's $\tau$ for all metric pairs.

\subsection{Abstractiveness}
Please see Fig.~\ref{fig:tac_abs_all_pairs},~\ref{fig:cnndm_abs_all_pairs} for Abstractiveness vs Kendall's $\tau$ for all metric pairs.
\begin{figure}
    \centering
    \foreach \x in {BScore_MS, BScore_R1, BScore_R2, BScore_RL, BScore_JS2, MS_R1, MS_R2, MS_RL, MS_JS2, R1_R2, R1_RL, R1_JS2, R2_RL, R2_JS2, RL_JS2}
    {
        \includegraphics[scale=0.2]{images/all_pairs_tac/\x_ref_abstractiveness_wrt_doc.png}
    }    
    \caption{Abstractiveness (x-axis) vs Kendall's $\tau$ (y-axis) for all metric pairs on the \texttt{TAC} dataset.}
    \label{fig:tac_abs_all_pairs}
\end{figure}
\begin{figure}
    \centering
    \foreach \x in {BScore_MS, BScore_R1, BScore_R2, BScore_RL, BScore_JS2, MS_R1, MS_R2, MS_RL, MS_JS2, R1_R2, R1_RL, R1_JS2, R2_RL, R2_JS2, RL_JS2}
    {
        \includegraphics[scale=0.2]{images/all_pairs_cnndm/\x_ref_abstractiveness_wrt_doc.png}
    }    
    \caption{Abstractiveness (x-axis) vs Kendall's $\tau$ (y-axis) for all metric pairs on the \texttt{CNNDM} dataset.}
    \label{fig:cnndm_abs_all_pairs}
\end{figure}

\subsection{Coverage}
\begin{figure}
    \centering
    \foreach \x in {BScore_MS, BScore_R1, BScore_R2, BScore_RL, BScore_JS2, MS_R1, MS_R2, MS_RL, MS_JS2, R1_R2, R1_RL, R1_JS2, R2_RL, R2_JS2, RL_JS2}
    {
        \includegraphics[scale=0.2]{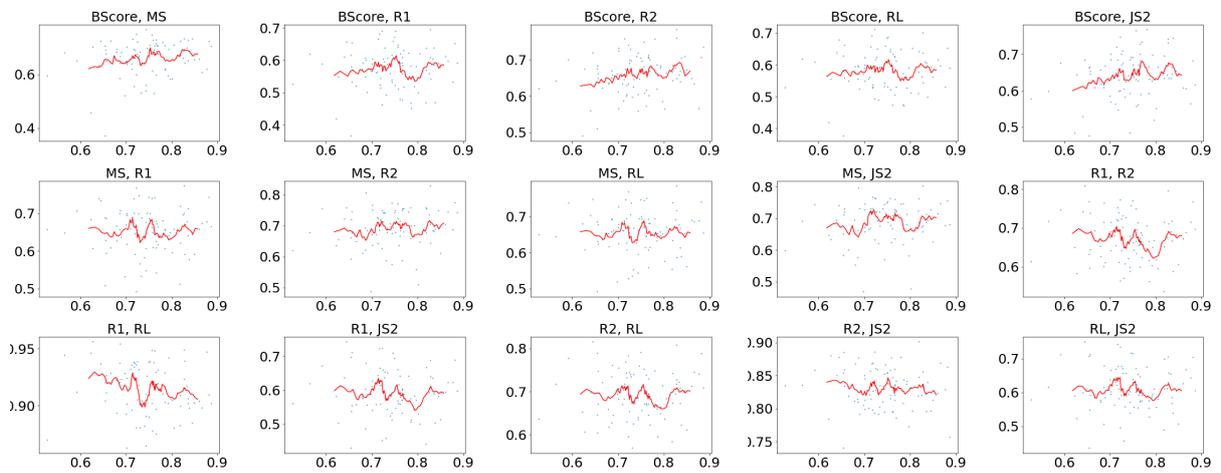}
    }    
    \caption{Coverage (x-axis) vs Kendall's $\tau$ (y-axis) for all metric pairs on the \texttt{TAC} dataset.}
    \label{fig:tac_cov_all_pairs}
\end{figure}

\begin{figure}
    \centering
    \foreach \x in {BScore_MS, BScore_R1, BScore_R2, BScore_RL, BScore_JS2, MS_R1, MS_R2, MS_RL, MS_JS2, R1_R2, R1_RL, R1_JS2, R2_RL, R2_JS2, RL_JS2}
    {
        \includegraphics[scale=0.2]{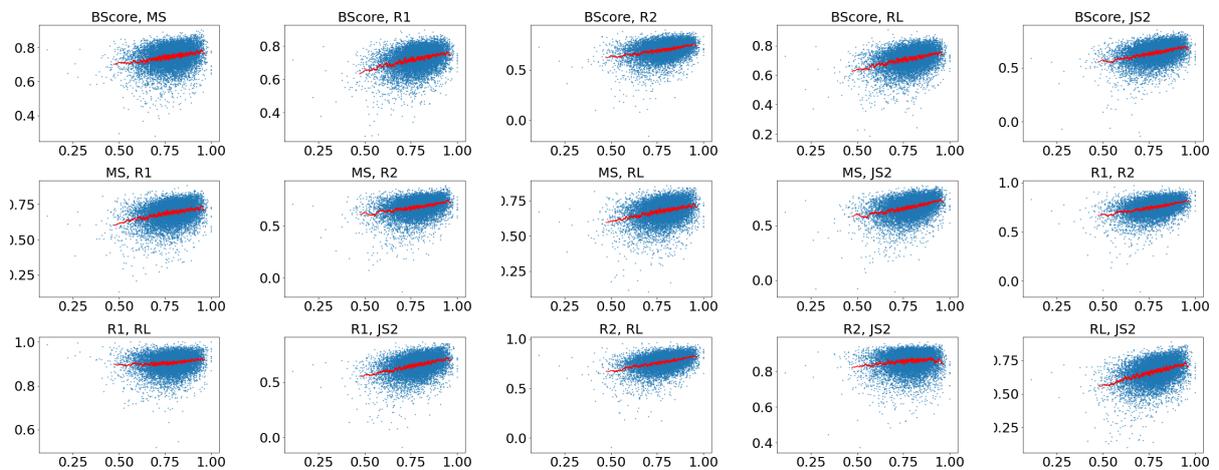}
    }    
    \caption{Coverage (x-axis) vs Kendall's $\tau$ (y-axis) for all metric pairs on the \texttt{CNNDM} dataset.}
    \label{fig:cnndm_cov_all_pairs}
\end{figure}

Please see Fig.~\ref{fig:tac_cov_all_pairs},~\ref{fig:cnndm_cov_all_pairs} for Coverage vs Kendall's $\tau$ for all metric pairs.

\end{document}